\documentclass[conference]{IEEEtran}
\IEEEoverridecommandlockouts

\usepackage{cite}
\usepackage{amsmath, amssymb, amsfonts}
\usepackage[dvipdfmx]{graphicx}
\usepackage{textcomp}
\usepackage{xcolor}
\usepackage{microtype}          
\usepackage[utf8]{inputenc}     
\usepackage[T1]{fontenc}
\usepackage{hyperref}           
\usepackage{url}
\usepackage{enumitem}           

\usepackage{color}

\usepackage{amsthm}
\usepackage{nicefrac}

\newcommand{\R}{\mathbb{R}}
\newcommand{\N}{\mathbb{N}}

\newcommand{\T}{\mathrm{T}}
\newcommand{\bs}{\boldsymbol}

\newcommand{\minimize}{\mathop{\rm minimize~~~~}\limits}
\newcommand{\subjectto}{\mathop{\rm subject~~to~~~~}}

\DeclareMathOperator*{\argmax}{argmax}
\DeclareMathOperator*{\argmin}{argmin}

\newcommand{\PS}{\mathrm{PS}}
\newcommand{\PF}{\mathrm{PF}}
\newcommand{\dominate}{\prec}
\newcommand{\dominateeq}{\preceq}

\usepackage[caption=false]{subfig}

\usepackage{booktabs}
\usepackage{multirow}
\usepackage{tabularx}

\usepackage{algorithm}
\usepackage{algorithmic}

\def\BibTeX{{\rm B\kern-.05em{\sc i\kern-.025em b}\kern-.08em
    T\kern-.1667em\lower.7ex\hbox{E}\kern-.125emX}}

\begin{document}

\title{Multi-start Optimization Method via Scalarization based on Target Point-based Tchebycheff Distance for Multi-objective Optimization}


\author{
    \IEEEauthorblockN{1\textsuperscript{st} Kota Nagakane}
    \IEEEauthorblockA{\textit{Institute of Science Tokyo} \\
    Yokohama, Japan \\
    nagakane.k@ic.comp.isct.ac.jp}
\and
    \IEEEauthorblockN{2\textsuperscript{nd} Masahiro Nomura}
    \IEEEauthorblockA{\textit{Institute of Science Tokyo} \\
    Yokohama, Japan \\
    nomura@comp.isct.ac.jp}
\and
    \IEEEauthorblockN{3\textsuperscript{rd} Isao Ono}
    \IEEEauthorblockA{\textit{Institute of Science Tokyo} \\
    Yokohama, Japan \\
    isao@comp.isct.ac.jp}
}


\maketitle

\IEEEpubidadjcol

\begin{abstract}
Multi-objective optimization is crucial in scientific and industrial applications where solutions must balance trade-offs among conflicting objectives.
State-of-the-art methods, such as NSGA-III and MOEA/D, can handle many objectives but struggle with coverage issues, particularly in cases involving inverted triangular Pareto fronts or strong nonlinearity.
Moreover, NSGA-III often relies on simulated binary crossover, which deteriorates in problems with variable dependencies.
In this study, we propose a novel multi-start optimization method that addresses these challenges.
Our approach introduces a newly introduced scalarization technique, the {\it Target Point-based Tchebycheff Distance} (TPTD) method, which significantly improves coverage on problems with inverted triangular Pareto fronts.
For efficient multi-start optimization, TPTD leverages a {\it target point} defined in the objective space, which plays a critical role in shaping the scalarized function.
This is because, if the target points are distributed uniformly, it is expected that single-objective function optimization using TPTD could construct an approximate solution set with good coverage.
The positions of the target points are adaptively determined according to the shape of the Pareto front, ensuring improvement in coverage.
The proposed method first searches for target points corresponding to objective vectors on the Pareto front boundary using a binary search method and then relocates the target points corresponding to objective vectors within the boundary according to the shape of the boundary.
This operation is computationally efficient because the positions of the non-boundary target points are determined in a single relocation.
Furthermore, the flexibility of this scalarization allows seamless integration with powerful single-objective optimization methods, such as natural evolution strategies, to efficiently handle variable dependencies.
Experimental results on benchmark problems, including those with inverted triangular Pareto fronts, demonstrate that our method outperforms NSGA-II, NSGA-III, and MOEA/D-DE in terms of the Hypervolume indicator.
Notably, our approach achieves computational efficiency improvements of up to 474 times over these baselines.
\end{abstract}

\begin{IEEEkeywords}
Multi-objective optimization, Scalarization method, Decomposition-based approach
\end{IEEEkeywords}

\section{Introduction}
\label{sec:introduction}

The multi-objective optimization problem (MOP) is an important problem faced in various fields of science and industry.
The objective of MOP is to find the Pareto solution set of multiple conflicting objective functions.
The image of the Pareto solution set in the objective space is called the Pareto front.
In general, if the objective function is a continuous function, the Pareto optimal solution set is an infinite set, so the purpose of the multi-objective optimization method is to find a finite approximate solution set that approximates the Pareto solution set.
The approximate solution set is evaluated by convergence and coverage \cite{pereira2022review}.
Many evolutionary computation methods have been proposed for multi-objective optimization problems \cite{deb2002fast, zitzler2001spea2, deb2013evolutionary, zhang2007moea, li2008multiobjective, panichella2022improved, sharma2022comprehensive, liang2022survey}.

NSGA-II \cite{deb2002fast} and SPEA2 \cite{zitzler2001spea2} are dominance-based methods that construct an approximate solution set by comparing solutions with each other using dominance relationships and performing survival selection \cite{li2015many, ishibuchi2008evolutionary}.
However, the performance of NSGA-II and SPEA2 deteriorate in terms of convergence as the number of objectives increases \cite{li2014evolutionary}.

NSGA-III \cite{deb2013evolutionary}, MOEA/D \cite{zhang2007moea}, and MOEA/D-DE\cite{li2008multiobjective} are the decomposition-based methods showing good performance in terms of convergence \cite{deb2013evolutionary, trivedi2016survey}.
NSGA-III utilizes reference points to partition the objective space while MOEA/D and MOEA/D-DE employ weight vectors and scalarization method.
They select the solutions whose objective vectors are closest to each partitioned region \cite{ishibuchi2019regular}.
They reportedly found an excellent approximate solution set in problems with a regular Pareto front \cite{ishibuchi2019regular}.
On the other hand, they have a problem in that the coverage of the approximate solution set degrades in problems with an inverted triangular Pareto front \cite{ishibuchi2019regular, he2020another}.
In MOEA/D, when the weight vectors are not consistent with the shape of the Pareto front, the solutions tend to be biased toward the boundaries of the Pareto front \cite{he2020another}. A similar issue is expected to occur in NSGA-III as well because the same idea improving the performance of MOEA/D for inverted triangular Pareto front has reportedly also improved that of NSGA-III \cite{he2020another}.
In addition, the performance of the Simulated Binary Crossover (SBX) \cite{deb1995simulated} often used in NSGA-II, SPEA2, and NSGA-III deteriorates in problems with variable dependencies~\cite{pan2021adaptive}.

In this paper, we propose a multi-start optimization method based on a new scalarization method named the Target Point-based Tchebycheff Distance method (TPTD), that is a decomposition-based method, to address the problems of the conventional methods.
TPTD uses the Tchebycheff distance between the objective vector of a solution and a target point in a target point set in the normalized objective space.
The target point set is formed to maximize coverage according to the shape of the Pareto front.
Additionally, the proposed method incorporates a Natural Evolution Strategy (NES) \cite{wierstra2014natural} that addresses variable dependencies, enhancing convergence in difficult optimization problems.
To assess the effectiveness of the proposed method, we compare the performance of the proposed method and that of NSGA-II, NSGA-III, and MOEA/D-DE using benchmark problems with decision vector spaces of 40 dimensions and objective spaces ranging from 3 to 5 dimensions, featuring both regular and inverted triangular Pareto fronts across linear, convex, and concave shapes.

\section{Problem Definition}
\label{sec:problem definition}

\newcommand{\SecProblemDefinitionEqOne}{
\begin{align}
    \label{eq:multi-objective function optimization problem}
    \begin{aligned}
        \minimize & \bs{f}(\bs{x}) = (f_1(\bs{x}), f_2(\bs{x}), \dots, f_m(\bs{x}))^\T, \\
        \subjectto & \bs{x} \in \Omega,
    \end{aligned}
\end{align}
}
\newcommand{\SecProblemDefinitionEqTwo}{
\begin{align}
    \label{eq:dominance and strong dominance relation}
    \begin{aligned}
        \bs{x} \dominateeq \bs{y} & \Leftrightarrow \forall i, f_{i}(\bs{x}) \leq f_{i}(\bs{y}) \land \exists j, f_{j}(\bs{x}) < f_{j}(\bs{y}),\\
        \bs{x} \dominate \bs{y} & \Leftrightarrow \forall i, f_{i}(\bs{x}) < f_{i}(\bs{y}).
    \end{aligned}
\end{align}
}
\newcommand{\SecProblemDefinitionEqThree}{
\begin{align}
    \label{eq:normalized multi-objective function optimization problem}
    \begin{aligned}
        \bs{f}'(\bs{x}) &= (f'_1(\bs{x}), f'_2(\bs{x}), \dots, f'_m(\bs{x}))^\T, \\
        f'_{i}(\bs{x}) &= \frac{f_{i}(\bs{x}) - z^{\text{ideal}}_{i}}{z^{\text{nadir}}_{i} - z^{\text{ideal}}_{i}}, \\
        z^{\text{nadir}}_{i} &= \max_{\bs{x} \in \PS} f_{i}(\bs{x}), ~ z^{\text{ideal}}_{i} = \min_{\bs{x} \in \PS} f_{i}(\bs{x}).
    \end{aligned}
\end{align}
}


The multi-objective optimization problem (MOP) is a class of problems that minimizes multiple objective functions simultaneously, formulated as follows: \SecProblemDefinitionEqOne
where $\bs{x}$ is a decision vector called solution, $\Omega \subseteq \R^{n}$ is a feasible region, $\bs{f}:\Omega \rightarrow \R^{m}$ is an objective function, and $f_{i}:\Omega \rightarrow \R$ is an $i$-th objective function.
The (strong) dominance relation \cite{deb2001multi} is used to compare solutions in MOPs.
Given two solutions $\bs{x}$ and $\bs{y}$, the (strong) dominance relation $\dominateeq$ ($\dominate$) is defined as follows: \SecProblemDefinitionEqTwo
$\bs{x}$ is called a (weak) Pareto optimal solution if there exists no solution $\bs{y} \in \Omega$ such that $\bs{y} \dominateeq (\dominate) \bs{x}$.
Generally, there are multiple Pareto optimal solutions, and the set of Pareto optimal solutions is called the Pareto optimal set, denoted as $\PS = \{\bs{x} \in \Omega | \nexists \bs{y} \in \Omega, \bs{y} \dominateeq \bs{x} \}$, and the set of their corresponding objective vectors is called the Pareto front, denoted as $\PF = \{ \bs{f}(\bs{x}) | \bs{x} \in \PS \}$.
The goal of multi-objective optimization is to find a finite approximate solution set that approximates the Pareto optimal set.
The quality of the approximate solution set is evaluated by convergence \cite{pereira2022review} and coverage \cite{pereira2022review}.
An objective function that is normalized so that the Pareto front is scaled within the range $[0,1]^{m}$ is called a normalized objective function, defined as follows: \SecProblemDefinitionEqThree

\section{Proposed Method}
\label{sec:proposed method}

\begin{figure}[tb]
    \centering
    \includegraphics[keepaspectratio, width=50mm]{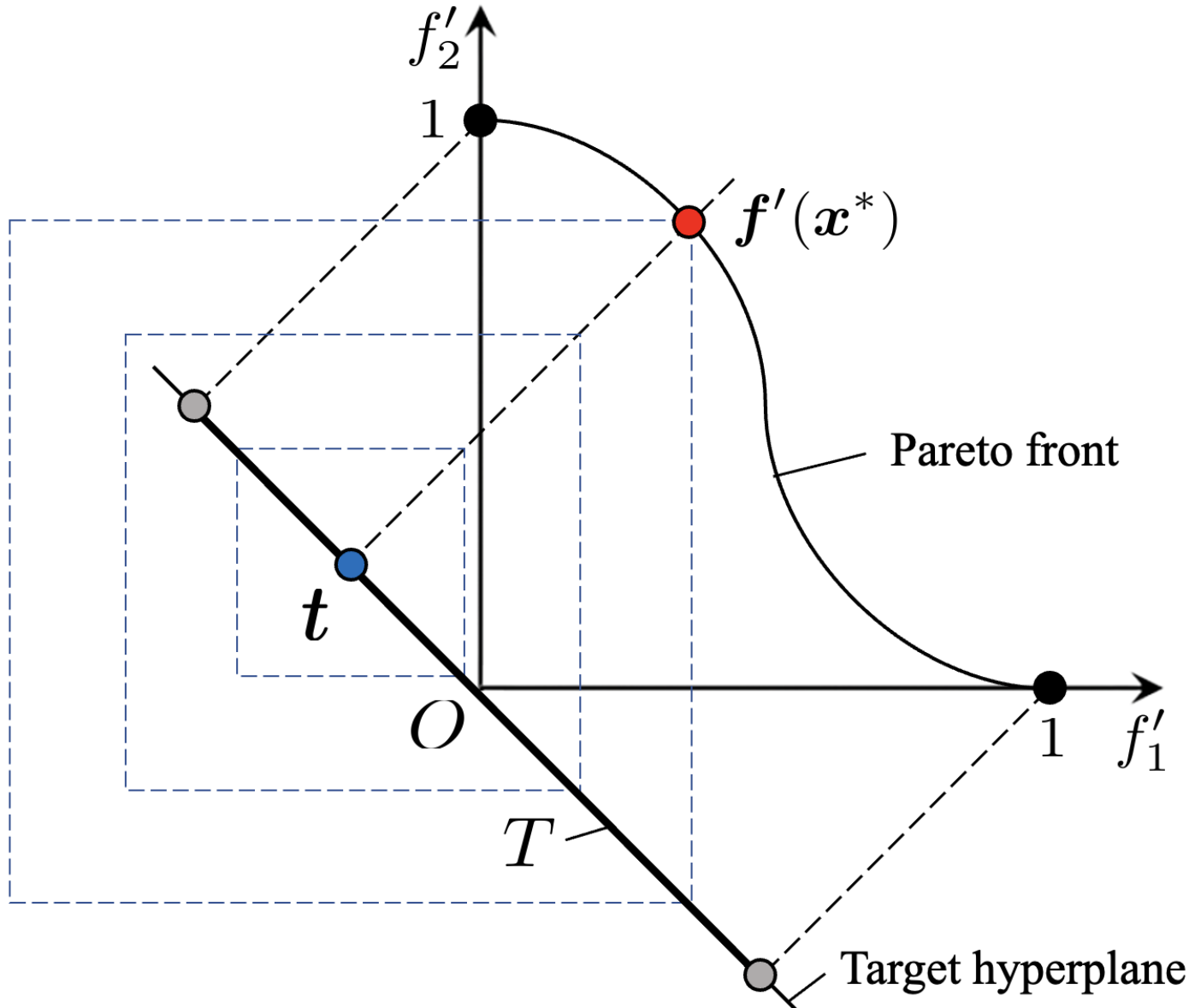}
    \caption{An example of the target point $\bs{t}$ and the normalized objective vector of optimal solution $\bs{f}'\left(\bs{x}^{*}\right)$ obtained by the Target Point-based Tchebycheff Distance method (TPTD) in a 2-objective problem. $T$ is the target point set. Note that the objective space is normalized.}
    \label{fig:overview of target point based tchebycheff distance}
\end{figure}

In this section, we propose a multi-start optimization method based on a new scalarization method named the Target Point-based Tchebycheff Distance (TPTD) method.
As shown in Fig.\ref{fig:overview of target point based tchebycheff distance}, TPTD utilizes the Tchebycheff distance between the objective vector $\bs{f}'\left(\bs{x}^{*}\right)$ of the solution $\bs{x}^{*}$ and the target point $\bs{t}$ on the $\left(m-1\right)$-dimensional target hyperplane in the normalized objective space.
The target points are placed on the target hyperplane to improve coverage according to the shape of the Pareto front.
To improve coverage, when a target point is given, it is desirable to quickly obtain its neighboring target points.
Thus, the proposed method employs {\it addresses} \cite{hamada2010adaptive}.
Addresses are vectors generated uniformly on an $\left(m-1\right)$-dimensional standard simplex, and the adjacency is managed by associating addresses with target points and their corresponding solutions.
The proposed method is a decomposition-based approach with a Natural Evolution Strategy (NES) \cite{wierstra2014natural} taking account of variable dependencies.
Therefore, it is expected to exhibit excellent search performance in many-objective problems with variable dependencies.

In the following, the details of TPTD is explained in \ref{subsec:target point based tchebycheff distance}.
In section \ref{subsec:search scenario of the proposed method}, we explain a search scenario, consisting of four steps, of the proposed method using TPTD.
Section \ref{subsec:step1}-\ref{subsec:step4} give the details of each step of the scenario, respectively.
Finally, we summarize the algorithm of the proposed method in Section \ref{subsec:algorithm}.

\subsection{Target Point based Tchebycheff Distance}
\label{subsec:target point based tchebycheff distance}

\newcommand{\SecImplicitSubsecTPTDEqOne}{
\begin{align}
    \label{eq:target point based tchebycheff distance}
    \begin{aligned}
        s^{\text{tptd}} \left(\bs{x} | \bs{f}', \bs{t}\right) = \max_{i \in \{1, \ldots, m\}} | f'_{i}\left(\bs{x}\right) - t_{i} |,
    \end{aligned}
\end{align}
}
\newcommand{\SecImplicitSubsecTPTDEqTwo}{
\begin{align}
    \label{eq:target point set}
    \begin{aligned}
        T &= \left\{ \bs{\pi} \left(\bs{u}\right) | \bs{u} = \left(u_{1}, \ldots, u_{m}\right)^{\T}, u_{i} \in [0, 1] \right\}, \\
        \bs{\pi} \left(\bs{u}\right) &= \bs{u} - \left( \frac{m - 2}{2 m} + \frac{1}{m} \sum_{i=1}^{m} u_{i} \right) \bs{1},
    \end{aligned}
\end{align}
}

\subsubsection{Definition of TPTD}

The Target Point-based Tchebycheff Distance (TPTD) method is defined in the normalized objective space by
\SecImplicitSubsecTPTDEqOne
where $\bs{x}$ is a solution vector, $\bs{f}'\left(\bs{x}\right)$ is a normalized objective vector of $\bs{x}$, $\bs{t} = \left(t_{1}, \ldots, t_{m}\right)^{\T} \in T$ is the target point, and $T$ is a target point set.
The target point set is a region projected onto the target hyperplane $\sum_{i=1}^{m} f'_{i} = - \frac{m - 2}{2}$ from the region $[0, 1]^{m}$ in the objective space, and is defined by \SecImplicitSubsecTPTDEqTwo
where $\bs{\pi}$ is a function projecting a vector in the objective space onto the target hyperplane, and $\bs{1}$ is an $m$-dimensional vector $\left(1, \ldots, 1\right)^{\T}$.

\subsubsection{Properties of the Optimal Solution of TPTD}
\label{subsubsec:properties of the optimal solution of TPTD}

\newcommand{\SecImplicitSubsecTPTDProofEqOne}{
\begin{align}
    \label{eq:single-objective optimization problem using TPTD}
    \begin{aligned}
        \minimize & s^{\text{tptd}} \left(\bs{x} | \bs{f}', \bs{t}\right) = \max_{i \in \{1, \ldots, m\}} | f'_{i}\left(\bs{x}\right) - t_{i} |, \\
        \subjectto & \bs{x} \in \Omega.
    \end{aligned}
\end{align}
}
\newcommand{\SecImplicitSubsecTPTDProofEqTwo}{
\begin{align}
    \label{eq:TPTD proof1}
    \begin{aligned}
        \bs{x}' \prec \bs{x}^{*} &\Leftrightarrow \forall i, f'_{i}\left(\bs{x}'\right) < f'_{i}\left(\bs{x}^{*}\right) \\
        &\Leftrightarrow \forall i, f'_{i}\left(\bs{x}'\right) - t_{i} < f'_{i}\left(\bs{x}^{*}\right) - t_{i}.
    \end{aligned}
\end{align}
}
\newcommand{\SecImplicitSubsecTPTDProofEqThree}{
\begin{align}
    \label{eq:TPTD proof2}
    \begin{aligned}
        \max_{i \in \{1, \ldots, m\}} |f'_{i}\left(\bs{x}\right) - t_{i}| = \max_{i \in \{1, \ldots, m\}} \left( f'_{i}\left(\bs{x}\right) - t_{i} \right),
    \end{aligned}
\end{align}
}
\newcommand{\SecImplicitSubsecTPTDProofEqFour}{
\begin{align}
    \label{eq:TPTD proof3}
    \begin{aligned}
        s^{\text{tptd}} \left(\bs{x}' | \bs{f}', \bs{t}\right) &= \max_{i \in \{1, \ldots, m\}} |f'_{i}\left(\bs{x}'\right) - t_{i}| \\
        & = \max_{i \in \{1, \ldots, m\}} \left( f'_{i}\left(\bs{x}'\right) - t_{i} \right) \\
        & < \max_{i \in \{1, \ldots, m\}} \left( f'_{i}\left(\bs{x}^{*}\right) - t_{i} \right) \\
        & \leq \max_{i \in \{1, \ldots, m\}} |f'_{i}\left(\bs{x}^{*}\right) - t_{i}| \\
        & = s^{\text{tptd}} \left(\bs{x}^{*} | \bs{f}', \bs{t}\right).
    \end{aligned}
\end{align}
}
\newcommand{\SecImplicitSubsecTPTDProofEqFive}{
\begin{align}
    \label{eq:TPTD proof4}
    \begin{aligned}
        f'_{k}\left(\bs{x}\right) - t_{k} &= f'_{k}\left(\bs{x}\right) - u_{k} + \frac{m - 2}{2m} + \frac{1}{m} \sum_{i=1}^{m} u_{i} \\
        &= f'_{k}\left(\bs{x}\right) - \frac{m - 1}{m} u_{k} + \frac{m - 2}{2m} + \frac{1}{m} \sum_{i \neq k} u_{i} \\
        & \ge f'_{k}\left(\bs{x}\right) - \frac{m - 1}{m} u_{k} + \frac{m - 2}{2m} + \frac{m - 1}{m} u_{k} \\
        &= f'_{k}\left(\bs{x}\right) + \frac{m - 2}{2m} \ge 0.
    \end{aligned}
\end{align}
}
\newcommand{\SecImplicitSubsecTPTDProofEqSix}{
\begin{align}
    \label{eq:TPTD proof5}
    \begin{aligned}
        & |f'_{k}\left(\bs{x}\right) - t_{k}| - |f'_{l}\left(\bs{x}\right) - t_{l}| \\
        & = f'_{k}\left(\bs{x}\right) + f'_{l}\left(\bs{x}\right) - u_{k} - u_{l} + \frac{m - 2}{m} + \frac{2}{m} \sum_{i=1}^{m} u_{i}  \\
        & = f'_{k}\left(\bs{x}\right) + f'_{l}\left(\bs{x}\right) - u_{k} - \frac{m - 2}{m} u_{l} + \frac{m - 2}{m} + \frac{2}{m} \sum_{i \neq l} u_{i}  \\
        & \ge f'_{k}\left(\bs{x}\right) + f'_{l}\left(\bs{x}\right) + \frac{m - 2}{m} u_{k} - \frac{m - 2}{m} u_{l} + \frac{m - 2}{m} \\
        & = f'_{k}\left(\bs{x}\right) + f'_{l}\left(\bs{x}\right) + \frac{m - 2}{m} \left(1 + u_{k} - u_{l}\right) \ge 0.
    \end{aligned}
\end{align}
}
\newcommand{\SecImplicitSubsecTPTDProofEqSeven}{
\begin{align}
    \label{eq:TPTD proof6}
    \begin{aligned}
        \max_{l \in L} |f'_{l}\left(\bs{x}\right) - t_{l}| & \leq |f'_{k}\left(\bs{x}\right) - t_{k}|.
    \end{aligned}
\end{align}
}
\newcommand{\SecImplicitSubsecTPTDProofEqEight}{
\begin{align}
    \label{eq:TPTD proof7}
    \begin{aligned}
        \max_{i \in \{1, \ldots, m\}} |f'_{i}\left(\bs{x}\right) - t_{i}| &= \max_{i \in \{1, \ldots, m\} \backslash L} |f'_{i}\left(\bs{x}\right) - t_{i}| \\
        &= \max_{i \in \{1, \ldots, m\} \backslash L} \left( f'_{i}\left(\bs{x}\right) - t_{i} \right) \\
        &= \max_{i \in \{1, \ldots, m\}} \left( f'_{i}\left(\bs{x}\right) - t_{i} \right).
    \end{aligned}
\end{align}
}

We prove by contradiction that the optimal solution $\bs{x}^{*}$ of the single-objective optimization problem (Eq.(\ref{eq:single-objective optimization problem using TPTD})) obtained by TPTD with the target point $\bs{t}$ on the target hyperplane is a weak Pareto optimal solution of the normalized multi-objective optimization problem $\bs{f}'$ where we assume $m \ge 2$.
\SecImplicitSubsecTPTDProofEqOne
Assuming that there exists a solution $\bs{x}' \in \Omega$ such that $\bs{x}' \prec \bs{x}^{*}$, we have: \SecImplicitSubsecTPTDProofEqTwo
If we can show that:
\SecImplicitSubsecTPTDProofEqThree
then
\SecImplicitSubsecTPTDProofEqFour
Thus, since $s^{\text{tptd}} \left(\bs{x}' | \bs{f}', \bs{t}\right) < s^{\text{tptd}} \left(\bs{x}^{*} | \bs{f}', \bs{t}\right)$, this contradicts the assumption that $\bs{x}^{*}$ is the optimal solution of Eq.(\ref{eq:single-objective optimization problem using TPTD}).
Therefore, $\bs{x}^{*}$ is a weak Pareto optimal solution of the normalized objective function $\bs{f}'$.
Hence, it is sufficient to prove Eq.(\ref{eq:TPTD proof2}).
For the target point $\bs{t} = \bs{u} - \left( \frac{m - 2}{2 m} + \frac{1}{m} \sum_{i=1}^{m} u_{i} \right) \bs{1}$, let $k = \argmin_{i \in \{1, \ldots, m\}} u_{i}$ be the number of the smallest element of $\bs{u}$.
Then, \SecImplicitSubsecTPTDProofEqFive
Let $L = \left\{ l \middle| l \in \{1, \ldots, m\}, f'_{l}\left(\bs{x}\right) - t_{l} < 0 \right\}$, then $k \not \in L$ and, for $l \in L$, \SecImplicitSubsecTPTDProofEqSix
Thus, \SecImplicitSubsecTPTDProofEqSeven
From this, \SecImplicitSubsecTPTDProofEqEight Therefore, Eq.(\ref{eq:TPTD proof2}) has been proved.

\subsection{Search Scenario of the Proposed Method}
\label{subsec:search scenario of the proposed method}

\begin{figure}[tb]
    \centering
    \includegraphics[keepaspectratio, width=90mm]{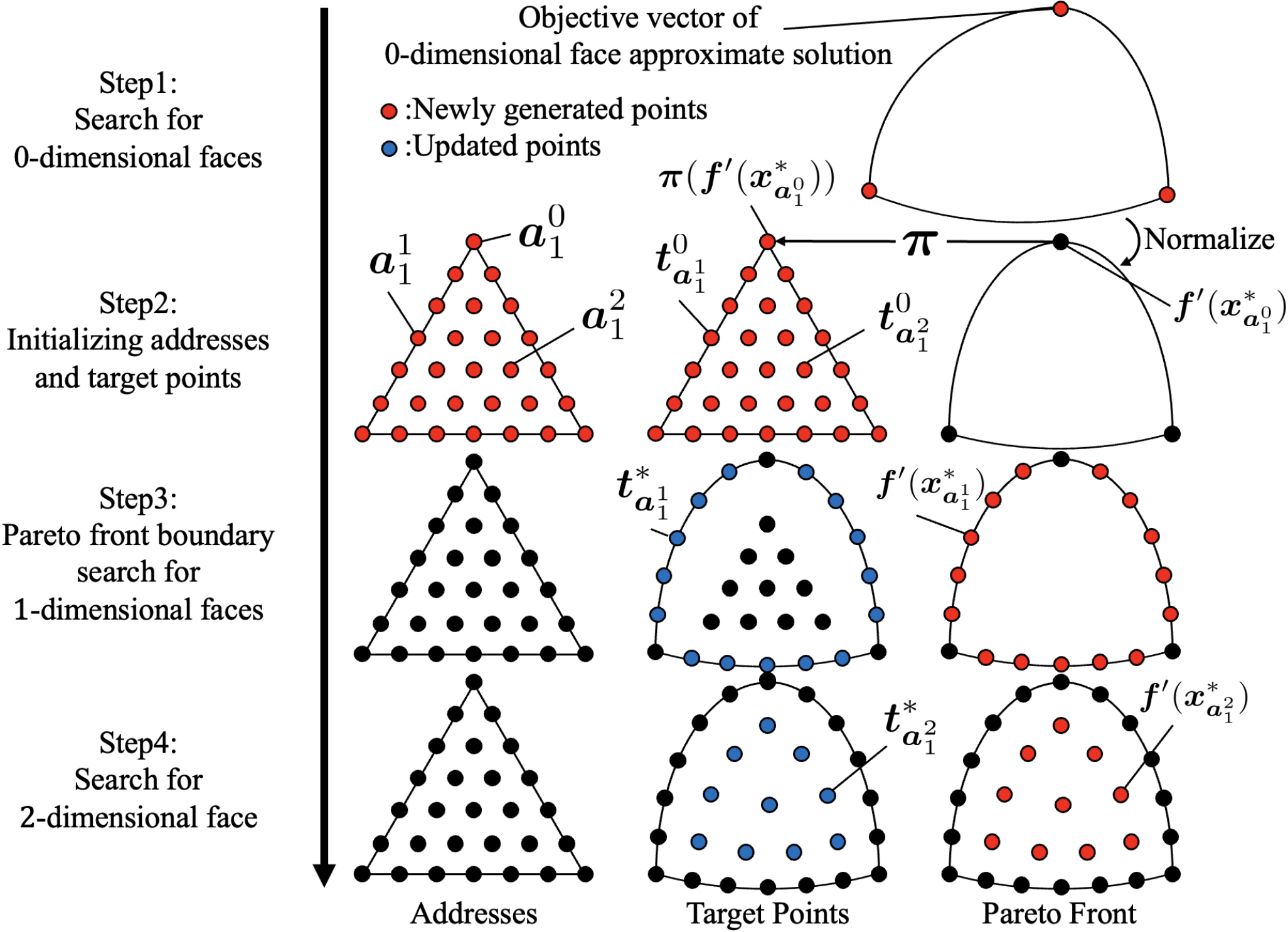}
    \caption{Search scenario of the proposed method in a 3-objective problem.}
    \label{fig:search scenario of proposed method}
\end{figure}

Figure.\ref{fig:search scenario of proposed method} shows a search scenario of the proposed method in a 3-objective problem ($m = 3$).

In Step 1, first, approximate solutions corresponding to the objective vectors of the $0$-dimensional face of the Pareto front are searched.
In this paper, vertices are called $0$-dimensional faces, and the faces of $n$-dimension are called $n$-dimensional faces \cite{hamada2010adaptive}.
Then, the normalized objective function $\bs{f}'$ is constructed using the maximum and minimum values of the objective vectors of the obtained approximate solutions.
Finally, the normalized objective vector set $F'^{*}_{0}$ of the obtained approximate solutions is constructed.
The details of Step 1 are explained in Section \ref{subsec:step1}.

In Step 2, addresses $\bs{a}_{i}^{d} \in A_d \,\, \left(d=0, 1, 2\right)$ and initial target points $ T_d^0 = \left \{ \bs{t}_{\bs{a}_{i}^{d}}^0 \middle| \bs{a}_{i}^{d} \in A_{d}, d=0, 1, 2 \right \}$ corresponding to the addresses are initialized, where the addresses $\bs{a}_{i}^{d} \in A_d \,\, \left(d=0, 1, 2\right)$ are uniformly generated on each face of the $2$-dimensional standard simplex, $d$ indicates the dimension of the face of the $2$-dimensional standard simplex, $i$ is the index of the address $\bs{a}_{i}^{d}$, $A_d$ is the address set of the $d$-dimensional face, and $\bs{t}_{\bs{a}_{i}^{d}}^0$ is an initial target point corresponding to the address $\bs{a}_{i}^{d}$. In this paper, $A_d$ and $T_d^0$ are called the {\it $d$-dimensional face address set} and the {\it $d$-dimensional face initial target point set}, respectively.
The initial target point $\bs{t}_{\bs{a}_{i}^{d}}^{0}$ is generated by affine transformation with the address $\bs{a}_{i}^{d}$, where $0$ indicates an initially generated point, and it is replaced with $*$ when the position is determined in Steps 3 and 4.
The affine transformation is performed so that the $0$-dimensional face address $\bs{a}_{i}^{0}$ matches the image of the normalized objective vector of the $0$-dimensional face approximate solution obtained in Step 1 on the target hyperplane $\bs{\pi}\left(\bs{f}'\left(\bs{x}^{*}_{\bs{a}_{i}^{0}} \right)\right)$, where$\bs{f}'\left(\bs{x}^{*}_{\bs{a}_{i}^{0}}\right) \in F'^{*}_{0}$.
This allows the initial target point set $T_d^0$ to be evenly distributed on the simplex spanned by the $\bs{\pi}\left(\bs{f}'\left(\bs{x}^{*}_{\bs{a}_{i}^{0}} \right)\right)$.
The details of Step 2 are explained in \ref{subsec:step2}.

In Step 3, the $1$-dimensional face approximate solution set $X_1^* = \left \{ \bs{x}_{\bs{a}_{i}^{1}}^{*} \middle| \bs{a}_{i}^{1} \in A_{1} \right \}$, the $1$-dimensional face normalized objective vector set $F'^{*}_{1} = \left \{ \bs{f}'\left(\bs{x}^{*}_{\bs{a}_{i}^{1}} \right) \middle| \bs{a}_{i}^{1} \in A_{1} \right \}$, and the $1$-dimensional face target point set $ T_1^* = \left \{ \bs{t}^{*}_{\bs{a}_{i}^{1}} \middle| \bs{a}_{i}^{1} \in A_{1} \right \}$ corresponding to the $1$-dimensional face address set $A_1$ are searched.
In this paper, this search is called {\it Pareto front boundary search} because $F'^{*}_{1}$ is on the boundary of the Pareto front.
In Step 3 of Fig.\ref{fig:search scenario of proposed method}, the normalized objective vector $\bs{f}'\left(\bs{x}^{*}_{\bs{a}_{1}^{1}} \right)$ and its corresponding target point $\bs{t}^{*}_{\bs{a}_{1}^{1}}$ corresponding to the initial target point $\bs{t}^{0}_{\bs{a}_{1}^{1}}$ are searched by the binary search based on NES with TPTD.
The details of Step 3 are explained in \ref{subsec:step3}.

In Step 4, after the $2$-dimensional face target point set $T_2^*$ is relocated to improve coverage according to $T_1^*$, i.e. the boundary of the Pareto front, the $2$-dimensional face approximate solution set $X_2^*$ and the $2$-dimensional face normalized objective vector set $F'^{*}_{2}$ are searched.
In Step 4 of Fig.\ref{fig:search scenario of proposed method}, $\bs{t}^{*}_{\bs{a}_{1}^{2}}$ is the relocated target point of $\bs{t}^{0}_{\bs{a}_{1}^{2}}$ in Step 2, and $\bs{f}'\left(\bs{x}^{*}_{\bs{a}_{1}^{2}} \right)$ is the normalized objective vector found by NES using TPTD with $\bs{t}^{*}_{\bs{a}_{1}^{2}}$.
The details of Step 4 are explained in \ref{subsec:step4}.

Note that the search for each approximate solution in Steps 3 and 4 can be executed independently for each target point, so acceleration by parallelization is expected.
Also note that, in 2-objective problems, Steps 3 and 4 are not executed and the approximate solution set $X_1^*$ is searched with the initial target points $T_1^0$.

\subsection{Step1: Search for 0-dimensional Faces}
\label{subsec:step1}

The proposed method utilizes the approximate solution set $X^{\text{tch}}$ obtained using the weighted Tchebycheff norm method\cite{miettinen1999nonlinear} as the $0$-dimensional face approximate solution set $X^{*}_{0}$ in the case of an inverted triangular Pareto front\cite{ishibuchi2019regular} while $X^{\text{mtch}}$ obtained using the modified Tchebycheff norm method\cite{shioda2015adaptive,deb2013evolutionary} in the case of a regular Pareto front\cite{ishibuchi2019regular}.
$X^{\text{tch}}$ and $X^{\text{mtch}}$ are defined by
\begin{align}
    \label{eq:step1 solution set by TCH and MTCH}
    \begin{aligned}
        \bs{w}_{i} &= (\underbrace{0, \dots, 0}_{i - 1}, 1, \underbrace{0, \dots, 0}_{m - i})^{\T}, ~ i \in \{1, \ldots, m\}, \\
        X^{\text{tch}} &= \left\{ \bs{x}^{\text{tch}}_{i} \middle| \bs{x}^{\text{tch}}_{i} = o_{\text{NES}} \circ s^{\text{tch}}\left(\bs{x} | \bs{f}, \bs{w}_{i}, \bs{z}^{\text{ideal}}\right) \right\}, \\
        X^{\text{mtch}} &= \left\{ \bs{x}^{\text{mtch}}_{i} \middle| \bs{x}^{\text{mtch}}_{i} = o_{\text{NES}} \circ s^{\text{mtch}}\left(\bs{x} | \bs{f}, \bs{w}_{i}, \bs{z}^{\text{ideal}}\right) \right\},\\
    \end{aligned}
\end{align}
where $\bs{w} = \left(w_{1}, \ldots, w_{m}\right)^{\T}$ is a weight vector, $s^{\text{tch}}$ and $s^{\text{mtch}}$ are Tchebycheff norm method and the modified Tchebycheff norm method, respectively, and $o_{\text{NES}} \circ s$ is the solution found by NES applied to the single objective function $s$.
The definitions of $s^{\text{tch}}$ and $s^{\text{mtch}}$ are as follows:
\begin{align}
    \label{eq:step1 TCH and MTCH}
    \begin{aligned}
        s^{\text{tch}}\left(\bs{x} | \bs{f}, \bs{w}, \bs{z}^{\text{ideal}}\right) & = \max_{i \in \{1, \ldots, m\}} w_{i} |f_{i}\left(\bs{x}\right) - z^{\text{ideal}}_{i}|, \\
        s^{\text{mtch}}\left(\bs{x} | \bs{f}, \bs{w}, \bs{z}^{\text{ideal}}\right) & = \max_{i \in \{1, \ldots, m\}} \frac{|f_{i}\left(\bs{x}\right) - z^{\text{ideal}}_{i}|}{w_{i}}.
    \end{aligned}
\end{align}
$\bs{z}^{\text{ideal}}$ is an ideal point given by
\begin{align}
    \label{eq:step1 ideal point}
    \begin{aligned}
        \bs{z}^{\text{ideal}} &= \left(z_{1}^{\text{ideal}}, z_{2}^{\text{ideal}}, \ldots, z_{m}^{\text{ideal}}\right)^{\T}, \\
        z^{\text{ideal}}_{i} &= f_{i}\left(\bs{x}^{*}\right), \\
        \bs{x}^{*} &= o_{\text{NES}} \circ f_{i}\left(\bs{x}\right), ~ i \in \{1, \ldots, m\}.
    \end{aligned}
\end{align}
If there is a solution in one solution set, $X^{\text{tch}}$ or $X^{\text{mtch}}$, that dominates a solution in the other solution set and the reverse does not hold, the solution set containing the dominant solution is chosen as the $0$-dimensional face approximate solution set $X_{0}^{*}$.
If there is no dominance relationship between the two solution sets, the solution set whose simplex volume $V\left(X\right)$ spanned by the objective vectors of the solution set $X \in \{X^{\text{tch}}, X^{\text{mtch}}\}$ is chosen as $X_{0}^{*}$.
The algorithm for $0$-dimensional face search is shown in Algorithm \ref{algorithm:Search 0-dimensional face}.
Additionally, the $i$-th normalized objective function is constructed using $X_{0}^{*}$ as follows:
\begin{align}
    \label{eq:step1 normalized objective function}
    \begin{aligned}
        f'_{i}\left(\bs{x}\right) &= \frac{f_{i}\left(\bs{x}\right) - f_{i}^{\text{min}}}{f_{i}^{\text{max}} - f_{i}^{\text{min}}}, \\
        f_{i}^{\text{max}} &= \max_{\bs{x} \in X_{0}^{*}} f_{i}\left(\bs{x}\right), ~
        f_{i}^{\text{min}} = \min_{\bs{x} \in X_{0}^{*}} f_{i}\left(\bs{x}\right).
    \end{aligned}
\end{align}

\begin{algorithm}[tb]
    \caption{Search$0$-dimensionalFaces}
    \label{algorithm:Search 0-dimensional face}
    \begin{algorithmic}[1]
        \REQUIRE $\bs{f}$, $o_{\text{NES}}$
        \ENSURE $X_{0}^{*}$

        \STATE Determine $\bs{z}^{\text{ideal}}$ according to Eq.(\ref{eq:step1 ideal point})
        \STATE Determine $X^{\text{tch}}, X^{\text{mtch}}$ according to Eq.(\ref{eq:step1 solution set by TCH and MTCH})
        \IF {$\exists \bs{x}^{\text{tch}}, \bs{x}^{\text{tch}} \dominateeq \bs{x}^{\text{mtch}} \land \nexists \bs{x}^{\text{tch}}, \bs{x}^{\text{mtch}} \dominateeq \bs{x}^{\text{tch}},$\\\hspace{15pt}$\bs{x}^{\text{tch}} \in X^{\text{tch}}, \bs{x}^{\text{mtch}} \in X^{\text{mtch}} $}
            \STATE $X_{0}^{*} = X^{\text{tch}}$
        \ELSIF {$\exists \bs{x}^{\text{mtch}}, \bs{x}^{\text{mtch}} \dominateeq \bs{x}^{\text{tch}} \land \nexists \bs{x}^{\text{mtch}}, \bs{x}^{\text{tch}} \dominateeq \bs{x}^{\text{mtch}},$\\\hspace{34pt}$\bs{x}^{\text{tch}} \in X^{\text{tch}}, \bs{x}^{\text{mtch}} \in X^{\text{mtch}} $}
            \STATE $X_{0}^{*} = X^{\text{mtch}}$
        \ELSE
            \STATE $X_{0}^{*} = \argmax_{X \in \{X^{\text{tch}}, X^{\text{mtch}}\}} V\left(X\right)$
        \ENDIF
    \end{algorithmic}
\end{algorithm}

\subsection{Step2: Initializing Addresses and Target Points}
\label{subsec:step2}

The address set $A$ is defined by
\begin{align}
    \label{eq:step2 address set}
    \begin{aligned}
        A &= \left\{ \bs{a} \middle|
            \begin{aligned}
                &\bs{a} \in [0, 1]^{m}, \bs{a} = \frac{1}{n_{\text{div}}} \bs{a}', \bs{a}' = \left(a'_{1}, \ldots, a'_{m}\right)^{\T}, \\
                &a'_{i} \in \N_{\ge 0}, \sum_{i=1}^{m} a'_{i} = n_{\text{div}},
            \end{aligned} \right\},
    \end{aligned}
\end{align}
where $n_{\text{div}}$ is a user parameter representing the number of divisions, and $\N_{\ge 0}$ is the set of non-negative integers.
The $d$-dimensional face address set $A_{d}$ is defined by
\begin{align}
    \label{eq:step2 d-dimensional address set}
    \begin{aligned}
        A_{d} &= \left\{ \bs{a} \in A \middle| \sum_{i = 1}^{m} \delta \left(a'_{i}\right) = m - n - 1 \right\}, \\
        \delta \left(a'_{i}\right) &= \left\{ \begin{aligned}
            &1, &a'_{i} = 0, \\
            &0, &\text{otherwise}.
        \end{aligned}
        \right.
    \end{aligned}
\end{align}
The initial target point set $T^{0}$ is constructed by using the address set $A$ and the affine matrix $\bs{B}$ as follows:
\begin{align}
    \label{eq:step2 initial target point}
    \begin{aligned}
        T^{0} &= \left\{ \bs{t}_{\bs{a}}^{0} | \bs{t}_{\bs{a}}^{0} = \bs{B} \bs{a}, \bs{a} \in A \right\}, \\
        \bs{B} &= \left( \bs{\pi}\left(\bs{f}'\left(\bs{x}_{\bs{a}^{0}_{1}}^{*}\right)\right), \ldots , \bs{\pi}\left(\bs{f}'\left(\bs{x}_{\bs{a}^{0}_{1}}^{*}\right)\right) \right),~ \bs{x}_{\bs{a}^{0}_{i}}^{*} \in X_{0}^{*}.
    \end{aligned}
\end{align}
The size of the address set and the initial target point set is given by $|A| = |T^{0}| = \frac{\left(n_{\text{div}} + m - 1\right)!}{n_{\text{div}}! \left(m-1\right)!}$.

\subsection{Step3: Pareto Front Boundary Search for 1 to (m-2)-dimensional Faces}
\label{subsec:step3}

\begin{figure}[tb]
    \centering
    \includegraphics[keepaspectratio, width=40mm]{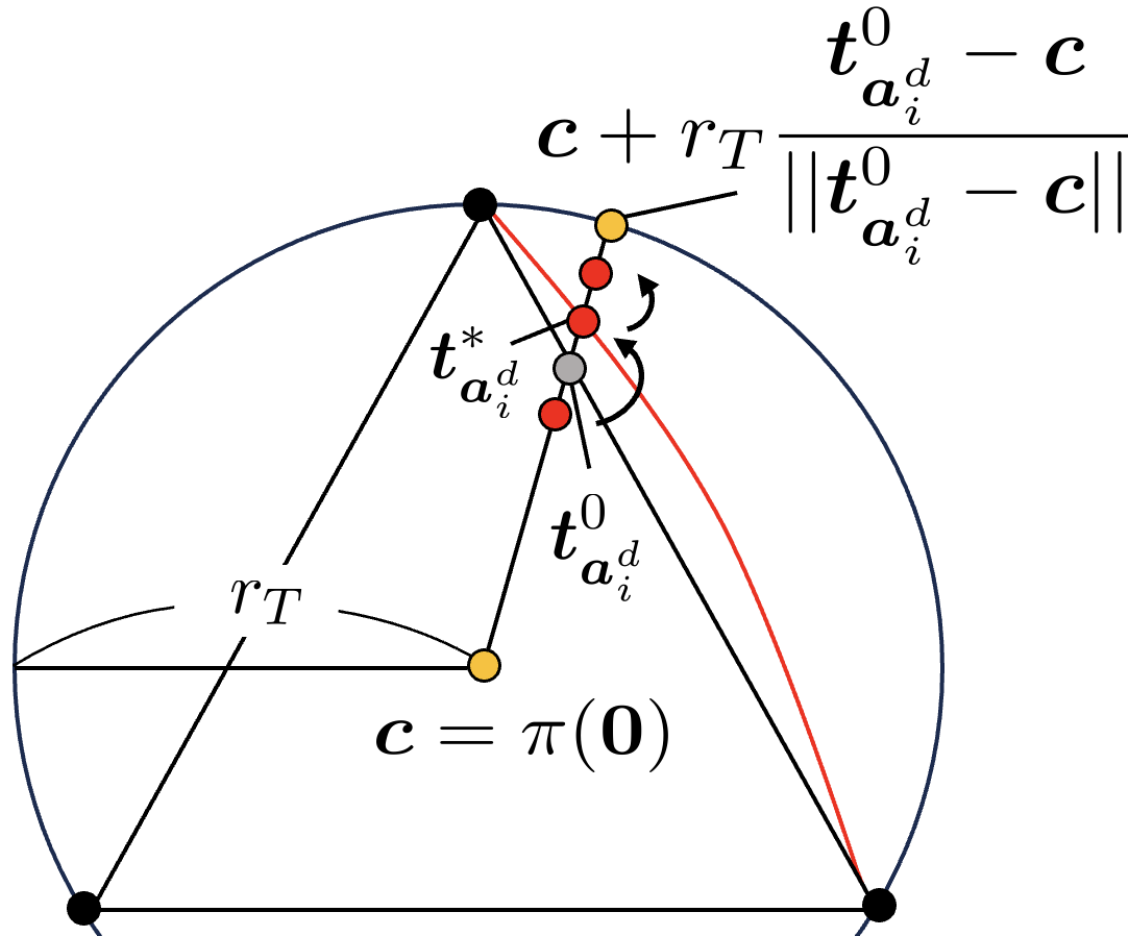}
    \caption{Pareto front boundary search in a 3-objective problem. The red line is the image of the Pareto front boundary on the target hyperplane, and the red points are the target points generated during the search.}
    \label{fig:pareto front boundary search}
\end{figure}

\begin{algorithm}[tb]
    \caption{ParetoFrontBoundarySearch}
    \label{algorithm:Pareto front boundary search}
    \begin{algorithmic}[1]
        \REQUIRE $\bs{f}'$, $o_{\text{NES}}$, $\bs{t}_{\bs{a}^{d}_{i}}^{0}$, $\epsilon_{t}$
        \ENSURE $\bs{t}^{*}_{\bs{a}^{d}_{i}}, \bs{x}^{*}_{\bs{a}^{d}_{i}}$

        \STATE Determine $\bs{t}^{\text{head}}$ and $\bs{t}^{\text{tail}}$ by Eq.(\ref{eq:step3 head and tail})
        \WHILE {}
            \STATE $\bs{t}^{\text{mid}} = \frac{1}{2} \left(\bs{t}^{\text{head}} + \bs{t}^{\text{tail}}\right)$
            \IF {$|| \bs{t}^{\text{head}} - \bs{t}^{\text{mid}} || < \epsilon_{t}$}
                \STATE {\bf break}
            \ENDIF
            \STATE $\bs{x}' = o_{\text{NES}} \circ s^{\text{tptd}}\left(\bs{x} | \bs{f}', \bs{t}^{\text{mid}}\right)$
            \IF {$\left| \left| \bs{f}'\left(\bs{x}'\right) - \bs{t}^{\text{mid}} - \frac{\sum_{i = 1}^{m} \left(f'_{i}\left(\bs{x}'\right) - t^{\text{mid}}_{i}\right) }{m} \bs{1}  \right| \right| \leq \epsilon_{t}$}
                \STATE $\bs{t}^{\text{head}} = \bs{t}^{\text{mid}}$
                \STATE $\bs{t}^{*}_{\bs{a}^{d}_{i}} = \bs{t}^{\text{mid}}$
                \STATE $\bs{x}^{*}_{\bs{a}^{d}_{i}} = \bs{x}'$
            \ELSE
                \STATE $\bs{t}^{\text{tail}} = \bs{t}^{\text{mid}}$
            \ENDIF
        \ENDWHILE

    \end{algorithmic}
\end{algorithm}

The Pareto front boundary search for $d = 1, \ldots, \left(m-2\right)$-dimensional faces employs the binary search method as shown in Fig.\ref{fig:pareto front boundary search}.
The algorithm of the Pareto front boundary search is shown in Algorithm \ref{algorithm:Pareto front boundary search}.
Line 1 initializes $\bs{t}^{\text{head}}$ and $\bs{t}^{\text{tail}}$.
\begin{align}
    \label{eq:step3 head and tail}
    \begin{aligned}
        \bs{t}^{\text{head}} &= \bs{c} = \pi \left(\bs{0}\right), \\
        \bs{t}^{\text{tail}} &= \bs{c} + r_{T} \frac{\bs{t}_{\bs{a}^{d}_{i}}^{0} - \bs{c}}{||\bs{t}_{\bs{a}^{d}_{i}}^{0} - \bs{c}||}, \\
        r_{T} &= \left\{ \begin{array}{cl}
            \frac{\sqrt{m}}{2} & \left(m \equiv 0 \mod 2\right) \\
            \frac{\sqrt{(m^{2} - 1)/m}}{2}  & \left(m \equiv 1 \mod 2\right) \\
        \end{array} \right.. \\
    \end{aligned}
\end{align}
Line 4 checks whether the distance between $\bs{t}^{\text{head}}$ and $\bs{t}^{\text{tail}}$ is less than a user parameter $\epsilon_\text{t}$.
Line 8 checks whether the objective vector $\bs{f}'\left(\bs{x}^{'} \right)$ is on the Pareto front by the following condition,
\begin{eqnarray}
    \label{eq:step3 target}
    \begin{aligned}
        \left| \left| \bs{f}'\left(\bs{x}'\right) - \bs{t}^{\text{mid}} - \frac{\sum_{i = 1}^{m} \left(f'_{i}\left(\bs{x}'\right) - t^{\text{mid}}_{i}\right) }{m} \bs{1}  \right| \right| \leq \epsilon_{t}.
    \end{aligned}
\end{eqnarray}

\subsection{Step 4: Search for (m-1)-dimensional Face}
\label{subsec:step4}

\begin{figure}[tb]
    \centering
    \includegraphics[keepaspectratio, width=60mm]{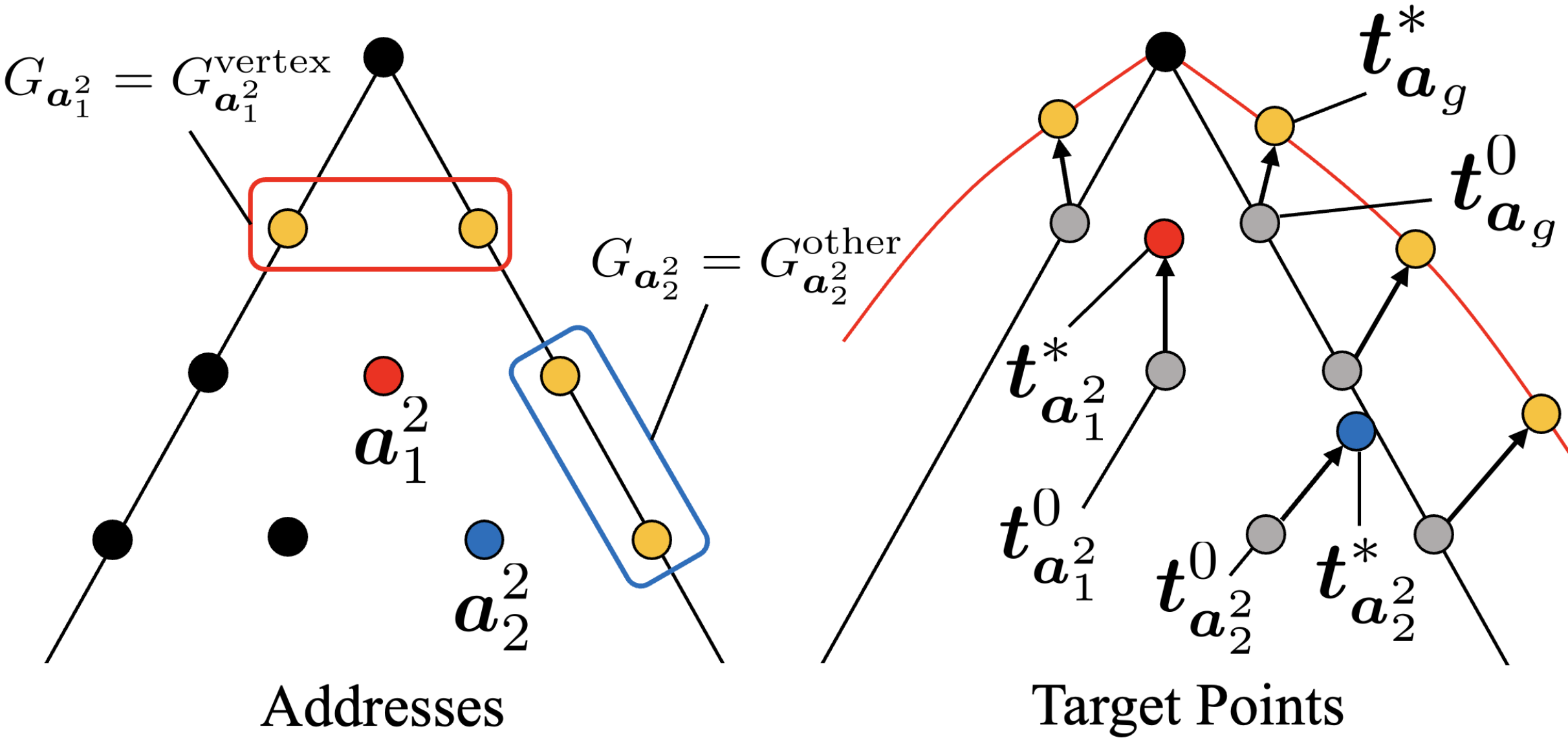}
    \caption{Example of guide point correspondence and relocation in a 3-objective problem. The red line is the image of the Pareto front boundary on the target hyperplane.}
    \label{fig:guide}
\end{figure}

First, as shown in Fig.\ref{fig:guide}, all the $(m-1)$-dimensional face target points $\bs{t}^{*}_{\bs{a}^{m-1}_{i}}$ in the $(m-1)$-dimensional face target point set $T_{m-1}^*$ are calculated by relocating $\bs{t}^{0}_{\bs{a}^{m-1}_{i}} \in T_{m-1}^0$ to improve coverage as follows:
\begin{align}
    \label{eq:guide point}
    \begin{aligned}
        \bs{t}^{*}_{\bs{a}^{m-1}_{i}} &= \bs{t}^{0}_{\bs{a}^{m-1}_{i}} + \eta \sum_{\bs{a}_{g} \in G_{\bs{a}^{m-1}_{i}}} \left( \bs{t}^{*}_{\bs{a}_{g}} - \bs{t}^{0}_{\bs{a}_{g}} \right).
    \end{aligned}
\end{align}
Where, $\eta$ is a user parameter for controlling the amount of movement and $G_{\bs{a}^{m-1}_{i}}$ is called the guide point set and defined by
\begin{align}
    \begin{aligned}
        G_{\bs{a}^{m-1}_{i}} &= \left\{
            \begin{aligned}
                & \emptyset, &\sum_{j=1}^{m} \zeta\left(a_{i,j}^{m-1}\right) = m, \\
                &G_{\bs{a}^{m-1}_{i}}^{\text{vertex}}, &\sum_{j=1}^{m} \zeta\left(a_{i,j}^{m-1}\right) \ge 2, \\
                &G_{\bs{a}^{m-1}_{i}}^{\text{other}}, &\text{otherwise}.
            \end{aligned}
        \right.\\
        \zeta\left(a_{i,j}^{m-1}\right) &= \left\{
            \begin{aligned}
                &1, &a_{i,j}^{m-1} = \min_{j \in \{1, \ldots, m\}} a_{i,j}^{m-1}, \\
                &0, &\text{otherwise},
            \end{aligned}
        \right. \\
        G_{\bs{a}^{m-1}_{i}}^{\text{vertex}} &= \left\{\bs{a}^{k} = \left(a^{k}_{1}, \ldots, a^{k}_{m}\right)^{\T} \middle| \begin{aligned}
            &k \in \{1, \ldots, m\} \backslash \{l\}, \\
            &l = \argmax_{j} a_{i,j}^{m-1}
        \end{aligned} \right\}, \\
        a^{k}_{j} &= \left\{
            \begin{aligned}
                &a_{j} + \frac{1}{n_{\text{div}}}, ~&j = l, \\
                &a_{j} - \frac{1}{n_{\text{div}}}, &j = k, \\
                &a_{j}, &\text{otherwise}.
            \end{aligned}
        \right.\\
        G_{\bs{a}^{m-1}_{i}}^{\text{other}} &= \left\{ \bs{a}^{q} = \left(a^{q}_{1}, \ldots, a^{q}_{m}\right)^{\T} \middle| \begin{aligned}
            &p \in \{1, \ldots, m\} \backslash \{q\},\\
            &q = \argmin_{j} a_{i,j}^{m-1}
        \end{aligned} \right\},\\
        a^{q}_{j} &= \left\{
            \begin{aligned}
                &a_{j} - \frac{1}{n_{\text{div}}}, ~&j = q, \\
                &a_{j} + \frac{1}{n_{\text{div}}}, &j = p, \\
                &a_{j}, &\text{otherwise}.
            \end{aligned}
        \right.
    \end{aligned}
\end{align}

Next, the $(m-1)$-dimensional face approximate solution set $X_{m-1}^*$ is searched by NES using TPTD with $T_{m-1}^*$.

\subsection{Algorithm}
\label{subsec:algorithm}


The algorithm of the proposed method is shown in Algorithm \ref{algorithm:proposed method}.
The proposed method takes the objective function $\bs{f}$, the NES $o_{\text{NES}}$, the number of divisions $n_{\text{div}}$, the parameter $\epsilon_\text{t}$ for determining whether the objective vector exists on the Pareto front, and the parameter $\eta$ for controlling the amount of movement in the relocation as input, and outputs the approximate solution set $X^{*}$.
Line 1 executes the $0$-dimensional face search in Step 1 and obtains the $0$-dimensional face approximate set $X_{0}^{*}$.
Line 2 constructs the normalized objective function $\bs{f}'$.
Lines 3-5 initialize the address set, $d$-dimensional face address set and the initial target point set in Step 2.
Line 6 initializes the approximate solution set $X^{*}$.
Lines 8-11 executes the $1$-dimensional face search with the initial target points if the number of objectives is $m = 2$.
Lines 13-29 execute Steps 3 and 4 if $m \ge 3$.
Line 13 initializes the set of determined target points $T^{*}$.
Lines 14-20 execute the Pareto front boundary search for $1, \ldots, (m-2)$-dimensional faces in Step 3.
Lines 21-29 execute the $(m-1)$-dimensional face search in Step 4.
Note that lines 8-11, 14-20, and 26-29 can be executed independently for each target point, and can be parallelized.

\begin{algorithm}[tb]
    \caption{ProposedMethod}
    \label{algorithm:proposed method}
    \begin{algorithmic}[1]
        \REQUIRE $\bs{f}$, $o_{\text{NES}}$, $n_{\text{div}}$, $\epsilon_{t}$, $\eta$
        \ENSURE $X^{*}$

        \STATE $X_{0}^{*} =$ Search$0$-dimensionalFaces$\left(\bs{f}, o_{\text{NES}}\right)$
        \STATE Determine $\bs{f}'$ according to Eq.(\ref{eq:step1 normalized objective function})

        \STATE Determine $A$ according to Eq.(\ref{eq:step2 address set})
        \STATE Determine $A_{d}, ~ d = 1, \ldots, m-1$ according to Eq.(\ref{eq:step2 d-dimensional address set})
        \STATE Determine $T^{0}$ according to Eq.(\ref{eq:step2 initial target point})

        \STATE $X^{*} = X_{0}^{*}$
        \IF {$m = 2$}
            \FOR {$\bs{a}^{1}_{i} \in A_{1}$}
                \STATE $\bs{x}^{*}_{\bs{a}^{1}_{i}} = o_{\text{NES}} \circ s^{\text{tptd}}\left(\bs{x} | \bs{f}', \bs{t}^{0}_{\bs{a}^{1}_{i}}\right)$
                \STATE $X^{*} = X^{*} \cup \{ \bs{x}^{*}_{\bs{a}^{1}_{i}} \}$
            \ENDFOR
        \ELSE
            \STATE $T^{*} = \emptyset$
            \FOR {$d = 1$ {\bf to} $m - 2$}
                \FOR {$\bs{a}^{d}_{i} \in A_{d}$}
                    \STATE $\bs{t}^{*}_{\bs{a}^{d}_{i}}, \bs{x}^{*}_{\bs{a}^{d}_{i}} =$ParetoFrontBoundary\\\hspace{60pt}Search$\left(\bs{f}', o_{\text{NES}}, \bs{t}^{0}_{\bs{a}^{d}_{i}}, \epsilon_{t}\right)$
                    \STATE $T^{*} = T^{*} \cup \{ \bs{t}^{*}_{\bs{a}^{d}_{i}} \}$
                    \STATE $X^{*} = X^{*} \cup \{ \bs{x}^{*}_{\bs{a}^{d}_{i}} \}$
                \ENDFOR
            \ENDFOR

            \FOR {$\bs{a}^{m-1}_{i} \in A_{m-1}$}
                \STATE Determine $G_{\bs{a}^{m-1}_{i}}$ according to Eq.(\ref{eq:guide point set})
                \STATE $\bs{t}^{*}_{\bs{a}^{m-1}_{i}} = \bs{t}^{0}_{\bs{a}^{m-1}_{i}} + \eta \sum_{\bs{a}_{g} \in G_{\bs{a}^{m-1}_{i}}} \left( \bs{t}^{*}_{\bs{a}_{g}} - \bs{t}^{0}_{\bs{a}_{g}} \right).$
                \STATE $T^{*} = T^{*} \cup \{ \bs{t}^{*}_{\bs{a}^{m-1}_{i}} \}$
            \ENDFOR
            \FOR {$\bs{a}^{m-1}_{i} \in A_{m-1}$}
                \STATE $\bs{x}^{*}_{\bs{a}^{m-1}_{i}} = o_{\text{NES}} \circ s^{\text{tptd}}\left(\bs{x} | \bs{f}', \bs{t}_{\bs{a}^{m-1}_{i}}^{*}\right)$
                \STATE $X^{*} = X^{*} \cup \{ \bs{x}^{*}_{\bs{a}^{m-1}_{i}} \}$
            \ENDFOR
        \ENDIF

    \end{algorithmic}
\end{algorithm}

\section{Experiments}
\label{sec:experiments and discussions}

\begin{table*}[tb]
    \centering
    \caption{Benchmark problems used in the performance comparison experiments.}
    \label{tab:problem}
    \begin{tabular}{|l|l|l|}
    \hline
    Problem & \multicolumn{2}{|l|}{Objective functions} \\ \hline \hline
    RP-Linear & \tiny $\begin{aligned} \left\{ \begin{aligned}
        f_{1}(\bs{x}) &= (1 + g(\bs{x}_{M})) \prod_{j=1}^{m-1} x_{j} \\
        f_{i}(\bs{x}) &= (1 + g(\bs{x}_{M})) (1 - x_{m-i+1}) \prod_{j=1}^{m-i} x_{j} \\
        f_{m}(\bs{x}) &= (1 + g(\bs{x}_{M})) (1 - x_{1})
    \end{aligned} \right. \end{aligned}$ &  \\ \cline{1-2}

    RP-Concave & \tiny $\begin{aligned} \left\{ \begin{aligned}
        f_{1}(\bs{x}) &= (1 + g(\bs{x}_{M})) \prod_{j=1}^{m-1} \sin \left(\frac{\pi}{2} x_{j} \right) \\
        f_{i}(\bs{x}) &= (1 + g(\bs{x}_{M})) \cos \left(\frac{\pi}{2} x_{m-i+1} \right) \prod_{j=1}^{m-i}  \sin \left(\frac{\pi}{2} x_{j} \right) \\
        f_{m}(\bs{x}) &= (1 + g(\bs{x}_{M})) \cos \left(\frac{\pi}{2} x_{1} \right)
    \end{aligned} \right. \end{aligned}$ & \tiny $\begin{aligned}
        g(\bs{x}_{M}) &= \sum_{i = m}^{n-1} \left\{ 100(x_{i+1} - x_{i}^{2})^{2} + (1- x_{i})^{2} \right\} \\
        \bs{x}_{M} &= (x_{m}, \ldots, x_{n})^{\T} \\
    \end{aligned}$ \\ \cline{1-2}

    RP-Convex & \tiny $\begin{aligned} \left\{ \begin{aligned}
        f_{1}(\bs{x}) &= (1 + g(\bs{x}_{M})) \prod_{j=1}^{m-1} \left(1- \sin \left(\frac{\pi}{2} x_{j} \right)\right) \\
        f_{i}(\bs{x}) &= (1 + g(\bs{x}_{M})) \left(1 - \cos \left(\frac{\pi}{2} x_{m-i+1} \right) \right) \prod_{j=1}^{m-i} \left(1 - \sin \left(\frac{\pi}{2} x_{j} \right) \right) \\
        f_{m}(\bs{x}) &= (1 + g(\bs{x}_{M})) \left(1- \cos \left(\frac{\pi}{2} x_{1} \right) \right)
    \end{aligned} \right. \end{aligned}$ & \\ \hline

    MED\cite{shioda2015adaptive} & \tiny $\begin{aligned}
        f_{i} (\bs{x}) = \left( \frac{1}{\sqrt{2}} \left|\left| \bs{x} - \bs{x}^{*}_{i} \right|\right| \right)^{p}
    \end{aligned}$ & \tiny $\begin{aligned}
        \bs{x}^{*}_{i} = (\underbrace{0, \dots, 0}_{i - 1}, 1, \underbrace{0, \dots, 0}_{n - i})^{\T}
    \end{aligned}$ \\ \hline
    \end{tabular}
\end{table*}

In this section, to confirm the effectiveness of the proposed method, we compare the performance of the proposed method, MOEA/D-DE \cite{li2008multiobjective}, NSGA-II \cite{deb2002fast} and NSGA-III \cite{deb2013evolutionary} using benchmark problems.

\subsection{Benchmark Problems}
\label{sec:benchmark problems}

We use $40$-dimensional and $3, 4, 5$-objective benchmark problems in Table.\ref{tab:problem}.
The RP-Linear problem, the RP-Concave problem, and the RP-Convex problem have a linear, concave, and convex regular Pareto front, respectively.
The RP problems have dependencies among parameters.
The RP problems are based on the DTLZ problem \cite{deb2005scalable}.
The MED problem has an inverted triangular Pareto front.
The parameter $p$ controls the convexity of the Pareto front. If $p < 1$, the Pareto front becomes concave. Otherwise, it becomes convex.
In this experiment, we set $p = 0.5, 1, 4$.

\subsection{Settings}

The user parameters of the proposed method are set to $n_{\text{div}} = 12$, $\epsilon_{t} = 0.01$, and $\eta = 0.4$.
CR-FM-NES \cite{nomura2022fast} is used as a natural evolution strategy.
For the MED problems, the population size is $10$, the maximum number of generations is $500$, and the initial step size is $0.5$.
For the RP problems, the population size is $40$, the maximum number of generations is $1500$, and the initial step size is $0.5$.
These parameters were determined by preliminary experiments.
The crossover method of MOEA/D-DE is rand/1/bin, and the scaling factor is $0.5$.
The mutation method is Polynomial Mutation (PM) \cite{deb1996combined}, and the mutation probability is $\frac{1}{n}$, and the distribution index is $20$.
The weight vectors used in MOEA/D-DE are generated uniformly on the $(m-1)$-dimensional standard simplex $\Delta^{m-1}$, as the addresses of the proposed method.
The crossover method of NSGA-II and NSGA-III is Simulated Binary Crossover (SBX) \cite{deb1995simulated}, and the distribution index is $20$.
The mutation method is PM.
We use jMetal \cite{durillo2011jmetal} to run the methods other than the proposed method.
With the $n_{\text{div}} = 12$, the size of the approximate solution set $|X|$ is $(m, |X|) = (3, 91), (4, 455), (5, 1820)$ for each number of objectives.
The number of trials is 30.

\subsection{Performance Indicators}

The Hypervolume (HV) \cite{zitzler1998multiobjective} and the execution time of the algorithm are utilized as performance indicators.
We use $\bs{r} = (1.1, \ldots, 1.1)^{\T}$ as the HV reference point.
The maximum execution time for each trial is set to 12 hours, and for methods that do not finish within 12 hours, the HV is calculated based on the approximate solution set at the 12-hour mark.

\subsection{Results and Discussions}

\begin{table*}[tb]
    \centering
    \caption{The average and standard deviation of the HV values of the approximate solution sets obtained by each method. The best HV value for each problem is shown in bold.}
    \label{tab:hv}
    \begin{tabular}{|c|c|c|c|c|c|}
    \hline
    Problem & $m$ & Proposed Method & MOEA/D-DE & NSGA-II & NSGA-III \\ \hline \hline
    & 3 & {\bf 0.82600 $\pm$ 0.00630} & 0.72094 $\pm$ 0.24442 & 0.00000 $\pm$ 0.00000 & 0.02982 $\pm$ 0.13985 \\ \cline{2-2}
    RP-Linear & 4 & {\bf 0.94401 $\pm$ 0.00213} & 0.90633 $\pm$ 0.00000 & 0.00000 $\pm$ 0.00000 & 0.00000 $\pm$ 0.00000 \\ \cline{2-2}
    & 5 & {\bf 0.98162 $\pm$ 0.00152} & 0.44121 $\pm$ 0.48600 & 0.00000 $\pm$ 0.00000 & 0.00000 $\pm$ 0.00000 \\ \hline

    & 3 & {\bf 0.53430 $\pm$ 0.00659} & 0.47149 $\pm$ 0.15985 & 0.00000 $\pm$ 0.00000 & 0.02184 $\pm$ 0.10921 \\ \cline{2-2}
    RP-Concave & 4 & {\bf 0.71354 $\pm$ 0.00603} & 0.66018 $\pm$ 0.00001 & 0.00000 $\pm$ 0.00000 & 0.00000 $\pm$ 0.00000 \\ \cline{2-2}
    & 5 & {\bf 0.83088 $\pm$ 0.00209} & 0.28581 $\pm$ 0.37208 & 0.00000 $\pm$ 0.00000 & 0.00000 $\pm$ 0.00000 \\ \hline

    & 3 & {\bf 0.97141 $\pm$ 0.00075} & 0.86902 $\pm$ 0.25880 & 0.00000 $\pm$ 0.00000 & 0.00485 $\pm$ 0.02132 \\ \cline{2-2}
    RP-Convex & 4 & {\bf 0.99333 $\pm$ 0.00048} & 0.95545 $\pm$ 0.18046 & 0.00000 $\pm$ 0.00000 & 0.00000 $\pm$ 0.00000 \\ \cline{2-2}
    & 5 & {\bf 0.99524 $\pm$ 0.00003} & 0.36823 $\pm$ 0.48926 & 0.00000 $\pm$ 0.00000 & 0.00000 $\pm$ 0.00000 \\ \hline

    & 3 & {\bf 0.09685 $\pm$ 0.00006} & 0.09219 $\pm$ 0.00011 & 0.09019 $\pm$ 0.00082 & 0.08243 $\pm$ 0.00228 \\ \cline{2-2}
    MED & 4 & {\bf 0.02932 $\pm$ 0.00001} & 0.02696 $\pm$ 0.00006 & 0.02705 $\pm$ 0.00020 & 0.02504 $\pm$ 0.00042 \\ \cline{2-2}
    ($p = 0.5$) & 5 & {\bf 0.00890 $\pm$ 0.00000} & 0.00732 $\pm$ 0.00002 & 0.00809 $\pm$ 0.00003 & 0.00673 $\pm$ 0.00282 \\ \hline

    & 3 & {\bf 0.28144 $\pm$ 0.00010} & 0.27292 $\pm$ 0.00011 & 0.25920 $\pm$ 0.00214 & 0.25427 $\pm$ 0.00275 \\ \cline{2-2}
    MED & 4 & {\bf 0.13718 $\pm$ 0.00006} & 0.12877 $\pm$ 0.00008 & 0.12627 $\pm$ 0.00056 & 0.11623 $\pm$ 0.02666 \\ \cline{2-2}
    ($p = 1$) & 5 & {\bf 0.06560 $\pm$ 0.00002} & 0.05268 $\pm$ 0.01468 & 0.05990 $\pm$ 0.00020 & 0.05306 $\pm$ 0.01500 \\ \hline

    & 3 & {\bf 0.94774 $\pm$ 0.00012} & 0.93541 $\pm$ 0.00008 & 0.92441 $\pm$ 0.00279 & 0.93737 $\pm$ 0.00184 \\ \cline{2-2}
    MED & 4 & {\bf 0.90446 $\pm$ 0.00014} & 0.88733 $\pm$ 0.00017 & 0.87867 $\pm$ 0.00214 & 0.88506 $\pm$ 0.00183 \\ \cline{2-2}
    ($p = 4$) & 5 & {\bf 0.85335 $\pm$ 0.00007} & 0.83211 $\pm$ 0.00015 & 0.82449 $\pm$ 0.00113 & 0.82320 $\pm$ 0.00226 \\ \hline
    \end{tabular}
\end{table*}

\begin{figure*}[tb]
    \centering
    \includegraphics[width=180mm]{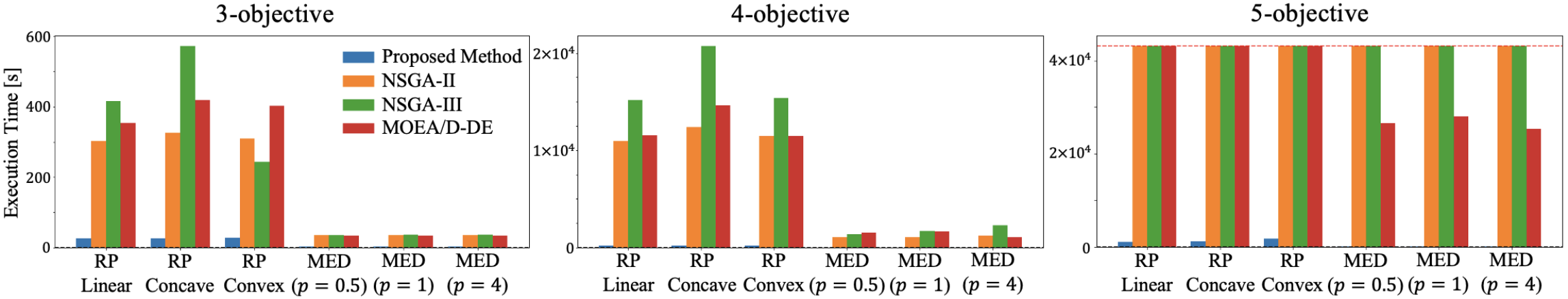}
    \caption{Execution time of each method on each problem. The red dashed line indicates the maximum execution time.
    The values of the proposed method (blue) are quite small, which may make them hard to see.
    }
    \label{fig:time}
\end{figure*}

Table.\ref{tab:hv} shows the average and standard deviation of the HV values of the approximate solution sets obtained by each method over 30 trials on each problem.
The best HV value for each problem is shown in bold.
As shown in Table.\ref{tab:hv}, the proposed method showed better performance than the others.
We believe that the search did not progress in NSGA-II and NSGA-III, which use SBX, because the RP problem has variable dependencies.
We believe that the performance of NSGA-III and MOEA/D-DE deteriorated in the MED problems because the MED problems have inverted triangular Pareto fronts.

Figure.\ref{fig:time} shows the average of execution time (seconds) of each method on each problem.
In Fig.\ref{fig:time}, the red dashed line indicates the maximum execution time.
As shown in Fig.\ref{fig:time}, the execution times of the proposed method are much shorter than those of the others on all the problems.
The speed up ratio is up to 474 times.
The computational complexity of the proposed method can be estimated based on the number of NES executions and the computational complexity of NES, ignoring the cost of evaluating the objective function.
The number of NES executions in the Pareto front boundary search can be estimated as $\log_{2} m$ because the range of the binary search depends on the number of objectives.
Therefore, the total number of NES executions in the proposed method is at most $N \log_{2} m$, where $N = |X|$ denotes the size of the approximate solution set.
On the other hand, assuming $n \geq m$, the computational complexity of one generation of CR-FM-NES can be estimated as $O(N_{\text{NES}} n)$ for the parameter update.
Here, $N_{\text{NES}}$ is the population size of NES.
Therefore, if the number of NES generations is denoted as $T_{\text{NES}}$, the computational complexity of the proposed method can be estimated as $O(T_{\text{NES}} N_{\text{NES}} N n \log_{2} m)$.
On the other hand, the computational complexity of NSGA-II is $O(T_{\text{EA}} N^2 m)$ \cite{deb2002fast}, the complexity of NSGA-III is $\max (O(T_{\text{EA}} N^2 \log^{m-2} N), O(T_{\text{EA}} N^2 m))$ \cite{deb2013evolutionary}, and the complexity of MOEA/D depends on the step that compares the scalarized function evaluation values of newly generated solutions with their neighbors, resulting in $O(T_{\text{EA}} N K m)$. Here, $T_{\text{EA}}$ denotes the number of generations in evolutionary algorithm, $K$ is the number of neighbors set in MOEA/D.
If the numbers of objective function evaluations in all the methods are equal, we have $T_{\text{EA}} = T_{\text{NES}} N_{\text{NES}} \log_{2} m$. Thus, the computational complexity of the proposed method can be rewritten as $O(T_{\text{EA}} N n)$.
From this, it can be seen that, when the size of the approximate solution set and the number of objective function evaluations in each method are equal to each other, respectively, the proposed method has a lower computational complexity than the other methods.
Note that, in this experiment, the proposed method was not parallelized, which means that the proposed method should be faster by parallelization.

\section{Conclusion}
\label{sec:conclusion}



In this paper, we proposed a multi-start optimization method based on a new scalarization method named the Target Point-based Tchebycheff Distance method (TPTD).
TPTD uses the Tchebycheff distance between the objective vector of a solution and a target point in a target point set in the normalized objective space.
The target point set is formed to maximize coverage according to the shape of the Pareto front.
Additionally, the proposed method incorporates a Natural Evolution Strategy (NES) that addresses variable dependencies, enhancing convergence in difficult optimization problems.
To assess the effectiveness of the proposed method, we compared the performance of the proposed method and that of NSGA-II, NSGA-III, and
MOEA/D-DE using benchmark problems with decision vector
spaces of 40 dimensions and objective spaces ranging from 3
to 5 dimensions, featuring both regular and inverted triangular
Pareto fronts across linear, convex, and concave shapes.
As the result, the proposed method outperformed NSGA-II, NSGA-III, and MOEA/D-DE in terms of the Hypervolume indicator and execution time. Notably, our approach achieved computational efficiency improvements of up to 474 times over these baselines.

For future work, we would like to parallelize the proposed method, reduce target points and approximate solutions for problems with a large number of objectives, and examine how the user parameters $\epsilon_{t}$ and $\eta$ affect its performance.
Furthermore, we plan to apply the proposed method to problems with irregular Pareto fronts, multi-objective discrete optimization problems, and real-world problems.

\bibliographystyle{IEEEtran}
\bibliography{ref}

\end{document}